\documentclass[11pt]{article}

\usepackage[final]{acl}
\usepackage{multirow}
\usepackage{hyperref}

\usepackage{times}
\usepackage{latexsym}
\usepackage{float}
\usepackage{listings}
\usepackage[T1]{fontenc}

\usepackage{booktabs}
\usepackage{svg}
\usepackage{url}

\usepackage{pdfpages}
\usepackage{cuted}
\usepackage{graphicx}

\usepackage{longtable}
\usepackage{array}
\usepackage[table]{xcolor}

\definecolor{conred}{HTML}{FF0000}
\definecolor{progreen}{HTML}{00B050}
 
\newcolumntype{L}{>{\raggedright\arraybackslash}p{2cm}}
\newcolumntype{M}{>{\raggedright\arraybackslash}p{15cm}}

\usepackage[utf8]{inputenc}

\usepackage{microtype}

\usepackage{inconsolata}

\usepackage{graphicx}

\usepackage{multirow}
\usepackage{makecell}

\usepackage{caption}

\usepackage{stfloats}

\title{Swiss4Arg: A  Parallel Dataset for Multilingual Argument Mining\\ across Swiss National Languages}

\author{
\textbf{Sophia Conrad}\textsuperscript{1*} 
\textbf{Fatma-Zohra Rezkellah}\textsuperscript{1*} 
\textbf{Christina Niklaus}\textsuperscript{2} 
\textbf{Reto Gubelmann}\textsuperscript{1} \\[0.5em]
\textsuperscript{1}Department of Computational Linguistics, University of Zurich \\
\textsuperscript{2}University of St. Gallen \\
\textsuperscript{*} These authors contributed equally to this work
}

\begin{document}
\maketitle

\begin{strip}
    \centering
    \vspace{-2cm}
    \includesvg[width=\textwidth]{final_images/teaser_figure}
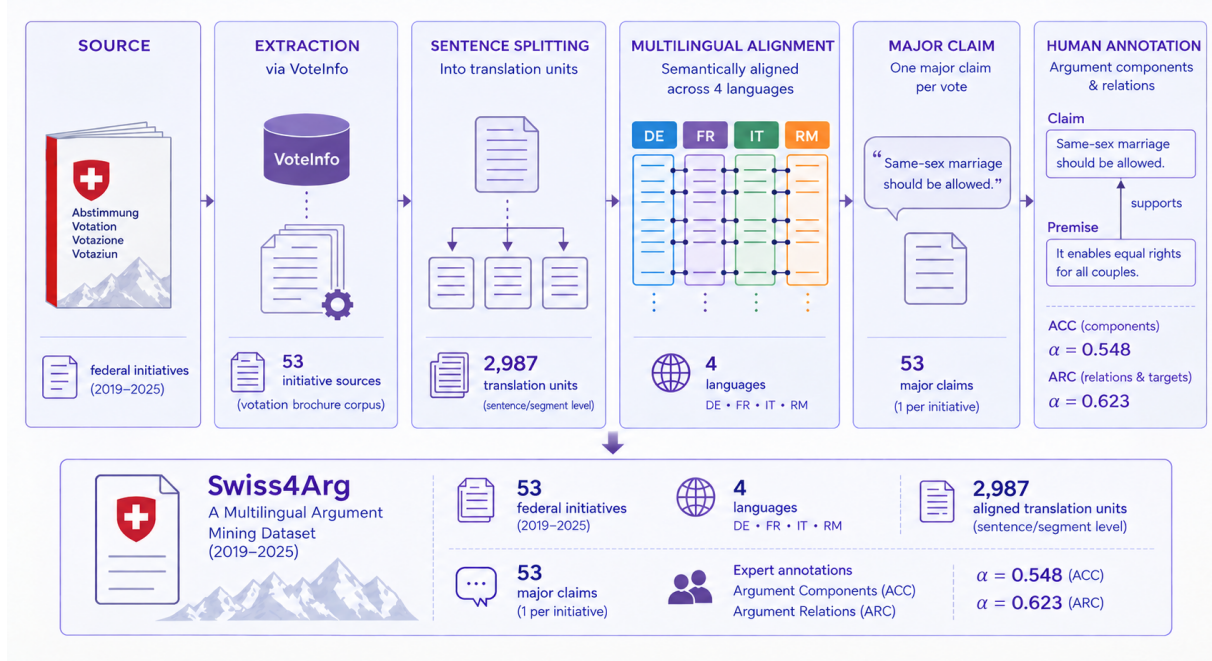
    \captionof{figure}{Creation pipeline of \texttt{Swiss4Arg}. The resulting dataset contains semantically aligned segments in all four languages, together with one major claim per vote and human annotations of argumentative components (ACC) and relations (ARC).}
    \label{fig:workflow}
\end{strip}

\begin{abstract} 



We present \texttt{Swiss4Arg}, a multilingual dataset for argument mining 
based on official Swiss voting brochures, covering all four national 
languages: German, French, Italian, and Romansh (a Romance minority language exclusively spoken in Switzerland).
The brochures contain explicitly formulated arguments systematically 
organized into pro and con sections, resulting in a cross-lingually 
aligned resource with balanced argumentative stance.
We create a dataset that comprises 53 federal voting issues collected between 2019 
and 2025, annotated for argument components (\textit{major claims}, 
\textit{claims}, \textit{premises}, and non-argumentative units) and 
relations (\textit{support} and \textit{attack}) by three native speakers 
per language. Inter-annotator agreement (Krippendorff's $\alpha$) reaches 0.548 for 
component classification and 0.623 for relation and target classification, 
computed jointly across all annotators on the parallel data.
We release both adjudicated labels and individual annotator annotations, 
enabling analyses of disagreement and cross-linguistic variation, and 
provide baseline results using GPT-5.4 and SwissBERT.
\texttt{Swiss4Arg} establishes a novel benchmark for multilingual argument 
mining, with a particular focus on cross-lingual comparability and 
low-resource settings such as Romansh. The dataset and code are publicly available at \href{https://anonymous-hf.up.railway.app/a/gozw0nion72y/}{HuggingFace} and \href{https://anonymous.4open.science/r/MultiVote-03BD/}{GitHub}.
\end{abstract}

\section{Introduction}
Switzerland provides a unique setting for multilingual argument mining (AM). 
It has four official languages: German, French, Italian, and Romansh, the 
last being a low-resource language with a small speaker base of 
ca.~38,000 as of 2024\footnote{\url{https://www.bfs.admin.ch/bfs/en/home/statistics/population/languages-religions/languages.html}}.
Switzerland's system of direct democracy requires that citizens regularly 
vote on federal issues, for which the government distributes official voting 
brochures presenting arguments both in favor of and against each proposal. 
Crucially, these materials are systematically produced in all four languages, 
resulting in parallel, high-quality texts with explicitly formulated and 
balanced argumentative content.

Multilingual AM resources remain limited, particularly for 
settings that combine multiple languages with aligned argumentative structure. 
Existing datasets are often restricted to a single high-resource language 
\cite[e.g.][]{stab-gurevych-2014-identifying,abkenar2025assessing, wambsganss-etal-2020-corpus}, leaving multilingual and low-resource AM underexplored.

In this paper, we present \texttt{Swiss4Arg} and make the following contributions:
\begin{itemize}\setlength\itemsep{0pt}\setlength\parskip{0pt}
    \item[(i)] A multilingual, cross-lingually aligned AM dataset based 
    on Swiss voting brochures, covering all four national languages including 
    Romansh.
    \item[(ii)] Fine-grained annotations of argument components (major claims, claims, 
    premises, and non-argumentative units) and relations (support and attack).
    \item[(iii)] Release of both individual annotator labels and adjudicated 
    (majority-vote) annotations, supporting studies of annotation variability 
    and disagreement.
    \item[(iv)] Baseline results for argument component and relation classification 
    using GPT-5.4 and SwissBERT.
\end{itemize}

\section{Related Work}
\paragraph{Swiss-languages Datasets.}
To the best of our knowledge, no AM corpus currently covers all 
four Swiss national languages. Notable multilingual Swiss resources include 
\texttt{SwiLTra-Bench} \citep{niklaus-etal-2025-swiltra}, covering all 
four languages, while most others focus on the three high-resource languages 
and exclude Romansh, including \texttt{X-Stance} \citep{vamvas2020xstance}, 
\texttt{SwissGov-RSD} \citep{wastl-etal-2026-swissgov}, and large-scale 
MT benchmarks \citep{deutsch-etal-2025-wmt24,andrews-etal-2025-bouquet}.
The most closely related resource is \texttt{VotingBooklets-Diamond} and 
its extension \texttt{VotingBooklets}, also derived from Swiss voting 
brochures, but focusing on document digitization and alignment rather than 
argumentative structure \citep{stussi-vamvas-2026-good}.
While these resources provide a strong multilingual foundation, none include 
annotations for argumentative structure, highlighting a clear gap that 
\texttt{Swiss4Arg} addresses.

\paragraph{Argument Framework.}
Argumentation has been studied through a variety of theoretical frameworks, 
including \citeposs{toulmin2003uses} model, \citeposs{walton2005argumentation} 
theory of argumentation schemes, and \citeposs{freeman2011dialectics} work on 
argument structure. While these approaches provide rich representations, they 
are difficult to operationalize consistently for large-scale corpus annotation. 
Consequently, most AM datasets adopt a simplified 
component--relation representation that is more suitable for reliable manual 
annotation and machine learning.

Following this tradition, we adopt the framework proposed by 
\citet{stab-gurevych-2014-annotating}, one of the most widely used schemes in argument 
mining. It represents argumentative structure using three component types 
(\textit{major claim}, \textit{claim}, \textit{premise}) and two relation types 
(\textit{support}, \textit{attack}). We choose this framework for its balance 
between expressive power, simplicity, and established use in prior work.
\paragraph{Monolingual Argument Datasets.}
Monolingual datasets are a core resource for training and evaluating AM systems, but appropriately annotated corpora remain limited, especially 
outside English \citep{wambsganss-etal-2020-corpus,lawrence-reed-2019-argument,fishcheva2019cross}. 
English remains the best-resourced language, with the persuasive essays corpus 
as a central benchmark, introduced with 90 and later extended to 402 essays
\citep{stab-gurevych-2014-identifying,stab_parsing_2017}, alongside the Argumentative 
Microtexts corpus and its extensions \citep{abkenar2025assessing,savigny2026amelia}. 
Resources for other languages are considerably fewer: German \citep{wambsganss-etal-2020-corpus}, 
Russian \citep{fishcheva2021traditional}, and recent work on Brazilian Portuguese, 
Chinese, and Spanish \citep{pereira2025evaluating,jahan2025winning}.
\paragraph{Multilingual Argument Mining.}
Multilingual AM is motivated by the scarcity of argument-annotated corpora 
outside English \citep{eger-etal-2018-cross,toledo-ronen-etal-2020-multilingual,
fishcheva2019cross,nabhani-khatib-2025-transfer}. Research has largely focused on transfer 
learning, either via data transfer (translating or projecting English annotations) 
or model transfer (fine-tuning multilingual models). Data transfer consistently 
outperforms model transfer for fine-grained component extraction, while zero-shot 
cross-lingual transfer remains less effective for complex span-level tasks \citep{eger-etal-2018-cross,yeginbergen2024crosslingual,yeginbergen_argument_2024,
jahan2025winning}.
\section{Dataset Creation}
Figure~\ref{fig:workflow} illustrates the complete workflow underlying our dataset,
from the extraction of official voting brochures to the final adjudicated annotations.
We describe each step in detail below.

\subsection{Data Collection \& Preparation}
\paragraph{Extracting the Data.}
Using an API call we extracted 53 federal voting texts spanning from February 2019 to November 
2025. A representative excerpt of a voting brochure, illustrating the typical structure of the summary, pro, and con sections, is provided in Appendix \ref{app:text}.

For our analysis, we focus on the \textit{Why yes} (pro) and \textit{Why no} (con) argument sections, which 
contain explicitly formulated argumentative content, and extract the 
\textit{short summaries} to retain contextual information for each voting 
question.
\paragraph{Aligning the Data.}
After sentence splitting the pro and con sections automatically using a regex pattern for 
Romansh and the NLTK  sentence tokenizer \citep{bird-loper-2004-nltk} for the other languages, we manually aligned the texts across all four languages using a custom tool based on the alignment tool used in \citet{wastl-etal-2026-swissgov}. That meant combining sentences into segments by performing many-to-one alignment whenever the same content was distributed across a different number of sentences in one or more  languages (see Fig.~\ref{fig:alignment}). We did not change or split any sentences, but only merged them into larger units. In the aligned data, each unit consistently captured the same semantic content across all languages. This alignment is essential for our cross-lingual experiments, as it guarantees that annotations are directly comparable across languages. On average, resolving these discrepancies required merging 7 sentences per voting text.

\begin{figure*}[htb]
    \centering
    \includesvg[width=\textwidth]{final_images/alignment_example}
    \caption{Example of manual sentence alignment across languages: sentences are merged into segments so that each unit captures the same semantic content in all four languages (here, the German and Romansh sentences in row 2 are moved up to align with the Italian and French counterparts).}
    \label{fig:alignment}
\end{figure*}

\begin{figure*}[h!]
    \centering
    \includesvg[width=0.8\textwidth]{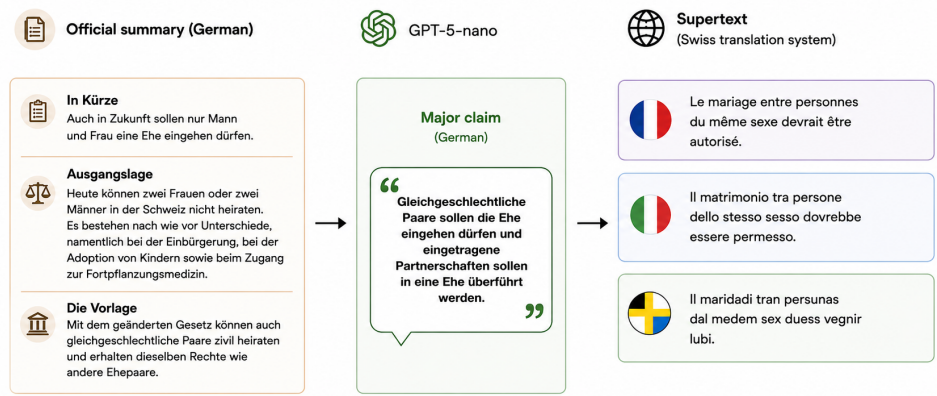}
    \caption{Illustration of the major claim generation process. The official German summary is condensed into a single synthetic major claim using GPT-5-nano. In this example, the generated German major claim (Same-sex marriage should be allowed) is translated into French, Italian, and Romansh using Supertext.}
    \label{major_claim}
\end{figure*}

\paragraph{Generating Major Claims.}
In \citeposs{stab-gurevych-2014-annotating} framework, the \textit{major claim} is a central component that expresses the author's overall stance toward the topic. In our dataset, however, the \textit{major claim} is not explicitly stated but implicitly conveyed through the summary section. To provide an explicit anchor for the argumentative structure, we generate a synthetic \textit{major claim} for each voting question by condensing the official summary into a single, concise statement that captures the core proposition under debate.
The \textit{major claims} were generated using GPT-5-nano  \citep{singh2025openai}, with the prompt provided in Appendix~\ref{claim_prompt}. Major claims were generated only in German, manually reviewed and, if necessary, revised by one of the authors. Twenty generated major claims (37.7\%) remained unchanged, while all others received minor revisions, primarily because the generated claims were either too vague or overemphasized less important details. The finalized major claims were subsequently translated into French, Italian, and Romansh using the commercial Swiss translation system Supertext\footnote{\url{https://www.supertext.com/en}}, which we 
preferred over general-purpose machine translation tools (e.g.\ 
Google Translate, DeepL) as it is specifically optimized for Swiss varieties of German, French, and Italian, and also supports Romansh (see Fig. \ref{major_claim}). Generating and validating the claims in a single language ensured that all languages shared the same underlying major claim.
\subsection{Annotations}
\paragraph{Annotators.}
We recruited three annotators per language, resulting in twelve annotators across the four languages, to perform two annotation tasks:
\vspace{-2mm}
\begin{enumerate}
    \item Argument component classification (ACC) into  \textit{claim}, and \textit{premise}, and \textit{non-argumentative} 
    \vspace{-2mm}
    \item Argument relation classification (ARC) into the three classes 
    \textit{neutral}, \textit{support}, and \textit{attack}
\end{enumerate}
\vspace{-1mm}
Annotators were native speakers of the language they annotated and university students enrolled in, at least, their third semester. They received a flat monetary compensation for their work. No prior annotation or NLP experience was required.

\paragraph{Pre-Annotation Workshop.}
The full annotation guidelines are available in the project repository at \href{https://anonymous.4open.science/r/MultiVote-03BD/}{GitHub}. 
They were written in English, with examples in the respective languages. To familiarize annotators with the guidelines and establish a shared understanding of the task, we conducted a one-hour online workshop attended by all annotators. The session was held in English to accommodate the multilingual group and covered 
the annotation scheme, data structure, and use of the annotation tool. The example discussed was an English translation of one voting sheet, allowing all participants to work with the same material regardless of their assigned language.
After the workshop, annotators completed an initial batch of annotations, which we reviewed for possible misunderstandings or recurring errors. We provided individual written feedback before full-scale annotation began. The most common comment towards annotators was to carefully distinguish \textit{non-argumentative} content and \textit{claims}. This calibration step helped resolve ambiguities early and improved overall consistency.

\subsection{Annotation \& Adjudication}
Annotations were collected using a custom HTML-based annotation tool 
that we developed specifically for this project\footnote{Available at \url{https://anonymous.4open.science/r/MultiVote-03BD/}.} and that enforced the core structural 
constraints of the annotation scheme. 
Because the dataset is parallel, adjudication was conducted jointly across all twelve annotators via majority voting. As shown in Table~\ref{tab:n_labels}, unanimous agreement was reached for 30.9\% of component labels and 67.0\% of relation labels, with majority vote resolving a further 66.5\% and  31.8\% respectively. The remaining 107 ties (2.5\% of components, 1.2\% of relations) were resolved by a designated adjudicator following the annotation guidelines. We release both the adjudicated labels and the full set of original annotations from all annotators to enable labeling disagreement analyses.

\begin{table*}[t]
\centering
\small
\setlength{\tabcolsep}{4pt}
\begin{tabular}{lrrrrrrrrr}
\toprule
 & \multicolumn{5}{c}{\textbf{Component}} & \multicolumn{4}{c}{\textbf{Relation}} \\
\cmidrule(lr){2-6} \cmidrule(lr){7-10}
 & \textbf{Claim} & \textbf{Premise} & \textbf{Non-Arg.} &  \textbf{Total} & \textbf{Percentage} & \textbf{Support} & \textbf{Attack} &  \textbf{Total} & \textbf{Percentage}\\
\midrule
Unanimous   & 368 & 307  & 168 & 843 & 29.3\%  & 1530 & 200 & 1730 & 67.0\% \\
Majority    & 398 & 1435 & 129 & 1962 & 68.1\% & 645  & 177 & 822 & 31.8\%  \\
Tie         & 22  & 53   & 1 & 76 & 2.6\%    & 21   & 10  & 31 & 1.2\%    \\
\midrule
Adjudicated & 793 & 1789 & 298 & 2881  &      & 2196 & 386 & 2583  &        \\
\bottomrule

\end{tabular}
\caption{Annotator agreement per label. The \textit{Adjudicated} row gives the final label counts after adjudication.}
\label{tab:n_labels}
\end{table*}


\section{Dataset Analysis}

\subsection{Stance Distribution}

The pro and con sections are of comparable length for each voting question, 
resulting in a balanced dataset with respect to argumentative stance. 
Across the 53 texts, there are a total of 1461 segments on the pro side 
and 1420 segments on the con side (see Fig.~\ref{fig:pro_con_distrib} in Appendix for the distribution of segments per text).

\subsection{Annotation Analysis}

\paragraph{Inter-Annotator Agreement.}
\label{sec:iaa}
Annotation consistency was evaluated using \citeposs{krippendorff2008systematic} $\alpha$, which accounts for chance agreement 
and supports multiple annotators.


Across all 12 annotators, we report $\alpha=0.548$ for component classification (\textit{claim} vs. \textit{premise} vs \textit{non-argumentative}, and $0.623$ for relation+target (\textit{support} vs. \textit{attack} towards a target component) classification which is comparable to \citeposs{hidey-etal-2017-analyzing} IAA. See Appendix~\ref{per_lang_disagreement} for agreement within languages.

\begin{figure}[h!]
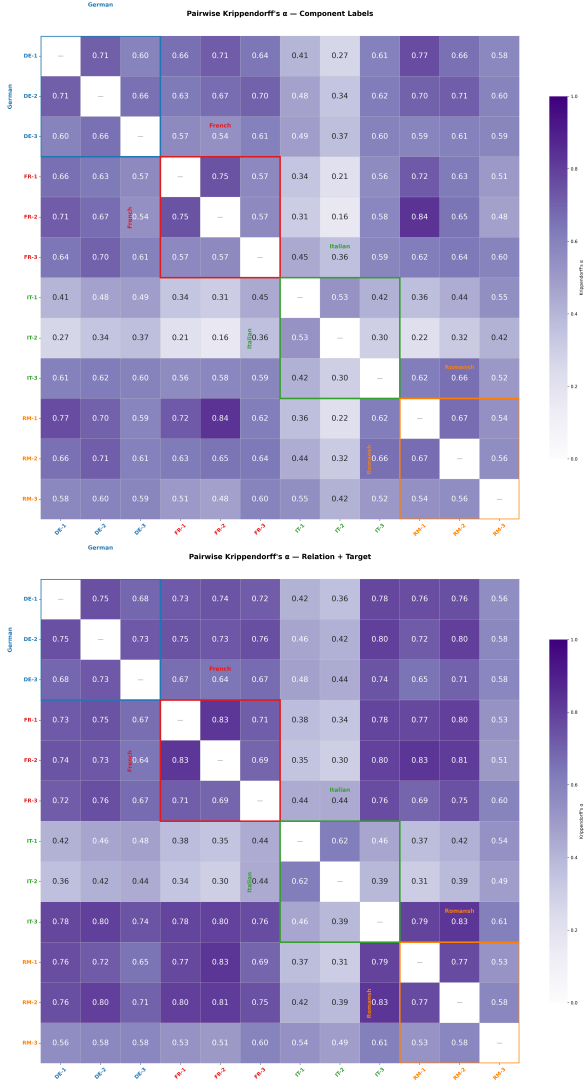

    \centering
    \includesvg[width=\linewidth]{final_images/pairwise_heatmap_component}
    \includesvg[width=\linewidth]{final_images/pairwise_heatmap_relation_target}
    \caption{Pairwise Krippendorff's $\alpha$ for component (top) and relation+target classification (bottom)
    across all 12 annotators.}
    \label{fig:pairwise_agreement}
\end{figure}

\paragraph{Pairwise Agreement and Cross-Lingual Consistency.}
To further investigate annotator agreement, we computed pairwise 
Krippendorff's $\alpha$ between all annotator pairs. Because the corpus 
consists of parallel voting brochures, annotators working in different 
languages labeled equivalent argumentative content, allowing meaningful 
cross-language comparisons. Figure~\ref{fig:pairwise_agreement} presents the resulting agreement matrices.

Component agreement ranges from 0.23 to 0.86 ($M = 0.58$), while 
relation agreement ranges from 0.30 to 0.83 ($M = 0.62$). 
A striking finding is that cross-lingual agreement between French and 
German annotators (ranging from 0.63 to 0.74 for components) is largely 
comparable to within-language agreement, and in some pairs exceeds it. 
This suggests that the argumentative structure of the brochures is 
sufficiently explicit and consistent across languages that annotators 
working independently on different language versions converge on similar 
labelings.

Italian annotators show a more complex picture. IT-1 exhibits consistently 
low agreement with all other annotators, including members of the Italian 
team (as low as 0.23 with FR-3 and 0.27 with FR-2). 
Romansh presents a different profile: within-language agreement is moderate 
(0.59--0.70), partly driven by RM-2, who shows weaker agreement with both within and outside of the Romansh team.

\paragraph{Disagreement Patterns.}
To characterize the nature of disagreements, 
~\ref{fig:conf_french}--\ref{fig:conf_romansh} in  Appendix~\ref{app:cms} show 
pairwise confusion matrices for component labels within each language. 
Across all languages, the dominant source of disagreement is the distinction 
between \textit{claims} and \textit{premises}, a well-documented challenge in AM
that often reflects ambiguity between intermediate conclusions and supporting 
evidence \citep{stab_parsing_2017}.
Confusions between argumentative and \textit{non-argumentative} spans are comparatively rare.

Inspection of the within-language confusion matrices reveals that IT-1 assigns more \textit{claim} labels compared to the other annotators, with 804 instances labeled 
\textit{claim} by IT-1 that IT-3 labeled differently, and 175 instances where IT-1 
used \textit{claim} while IT-3 used \textit{premise}. This is further reflected in the 
per-language annotator distributions (see Fig.~\ref{fig:per_lang_dist} in Appendix): 
Italian annotators as a group assign the \textit{claim} label at a markedly higher 
rate (42.0\%) compared to German (29.4\%), French (25.8\%), and Romansh 
(29.4\%), while assigning \textit{premise} at a substantially lower rate (35.8\% 
vs.\ approximately 55--60\% for the other languages). By contrast, IT-3 
shows strong cross-lingual agreement with French and German annotators 
(up to 0.80 with FR-3 in relation+target), suggesting that this divergence 
is driven primarily by IT-1. For Romansh, RM-2 tends to assigns the 
\textit{claim} label more frequently than the other two, with 601 instances labeled \textit{claim} by RM-2 that RM-3 labeled 
as \textit{premise}.

\subsection{Adjudicated Dataset}
\paragraph{Label Distribution.}
The bottom row of Table~\ref{tab:n_labels} presents the distribution of labels in the final adjudicated dataset. As expected from the data, \textit{support} and \textit{attack} relations are balanced for claims, but premises overwhelmingly support rather than attack their target claims, as each section argues only for its stance. 



\textit{Premises} constitute the largest category (60.1\%), followed by \textit{claims }
(26.4\%), consistent with the structure of the brochures in which each 
sub-position is typically supported by multiple pieces of evidence. 
\textit{Non-argumentative} units account for 10.0\% of all units, corresponding 
to section headers, procedural sentences, and contextual background. This shows the argumentative density of these texts, which is considerably higher than in other argumentative texts such as essays. 
The distribution is notably stable across pro and con sections, 
\textit{claims} appear at 26.2\% (Pro) and 26.6\% (Con), and \textit{premises} at 
59.7\% (Pro) and 60.5\% (Con).



\section{Baseline Argument Classification}
\label{sec:baselines}
To establish performance benchmarks on our dataset, we evaluate two baseline systems on the argument classification task: a large multilingual language model, GPT-5.4 \citep{singh2025openai}, and a domain-adapted Swiss multilingual encoder, SwissBERT \citep{vamvas-etal-2023-swissbert}. Both are evaluated against the majority-vote ground truth described above, on the exact two subtasks performed by annotators: (1)~component classification, and (2)~relation+target prediction.

\subsection{GPT-5.4}
\label{sec:gpt}
We query GPT-5.4 in a zero-shot setting using a structured prompt grounded 
in the theoretical frameworks of \citet{stab_parsing_2017} and  
\citet{peldszus2015joint}.
The system prompt specifies the component taxonomy and the strict 
argumentative hierarchy (premise $\to$ claim $\to$ major\_claim), while 
the user prompt supplies the document's major claim and all pre-segmented 
text units, each identified by a \texttt{segment\_id}, and requests a 
structured JSON response\footnote{Major Claims are pre-annotated and 
excluded from model evaluation.}
(see Appendix~\ref{app:prompts} for the full prompts).
Outputs are extracted from the JSON response and normalised to match the 
annotation schema; a retry mechanism is triggered when fewer than 80\% of 
segments receive a valid annotation. 
The model is evaluated independently on each of the four languages. 

\subsection{SwissBERT}\label{sec:swissbert}
In addition to querying GPT, we fine-tune SwissBERT, a transformer model with language adapters for each of the four Swiss languages, on the two tasks separately. Both tasks are class-imbalanced, so we use inverse-frequency class weighting during training. Given the small number of proposals (53), we use 5-fold cross-validation at the proposal level rather than a single split: proposals are partitioned into 5 folds, and for each fold $k$, fold $k$ serves as test, fold $(k+1)\ mod\ 5$ as validation, and the rest as training, so every proposal is tested and validated exactly once. Learning rate and epoch count are tuned by grid search on a single held-out fold, selecting by validation macro-F1, before running the full cross-validation with the chosen hyperparameters\footnote{See Appendix~\ref{app:swissbert_hyperparams}}. We report the same scores as for GPT-5.4 in Table~\ref{tab:component-comparison} and \ref{tab:gpt-relation} for direct comparison with the GPT model.

\paragraph{Component classification results.}\label{}
Table~\ref{tab:component-comparison} reports precision, recall, and F1 per class, 
as well as macro and weighted F1, for each language individually and averaged across languages.
GPT-5.4 achieves an overall macro F1 of 0.556 and accuracy of 0.601\%, 
with performance varying moderately across languages (macro F1 ranging 
from 0.528 for French to 0.582 for Romansh).
Notably, Romansh, the lowest-resource language in our corpus, yields the 
best component classification performance.

SwissBERT fine-tuned on \texttt{Swiss4Arg} is consistently outperforming zero-shot GPT-5.4 with an average Macro F1 of 0.859 and accuracy of 0.866. This once again shows how powerful BERT-family models still are, as already shown by e.g. \citet{bucher2024fine, zhang-etal-2025-bert-like} for text classification or \citet{djouvas2024llms} for argument strength classification.
Furthermore, SwissBERT's performance is comparable across all four languages, Macro F1 being in a small range from 0.853 to 0.866. 

Across all languages, performance is highly uneven across classes for the GPT model:
\textit{Premise} is identified most reliably (F1 $\approx$ 0.71), as it 
is the dominant class (60.1\% of items) and its supporting-evidence 
function is relatively unambiguous.
\textit{Claim} is substantially harder (F1 $\approx$ 0.43): the model 
achieves moderate recall but low precision.
This is confirmed by the confusion matrices (see Fig.~\ref{fig:gpt-confusion} in Appendix~\ref{app:gpt_cms}), 
which show that roughly 468--515 ground-truth \textit{premises} are predicted as 
\textit{claims} in each language.
\textit{Non-Argumentative} exhibits the inverse pattern: high recall 
(0.779--0.802) but low precision (0.355--0.435), meaning the model 
correctly catches most \textit{non-argumentative} segments but also mislabels a 
non-trivial number of argumentative ones as such.

In contrast, the per-class results are closer together in the SwissBERT predictions. F1 ranges from 0.774 for \textit{claims}, to 0.901 and 0.902 for \textit{premises} and non-argumentative units, respectively. While precision is consistently highest for \textit{non-argumentative} units, \textit{premises} are identified with the highest recall. Surprisingly, recall for \textit{claims} is quite low at ~0.70 for all languages.

\begin{table*}[h]
\centering
\footnotesize
\renewcommand{\arraystretch}{1.2}
\setlength{\tabcolsep}{7pt}
\begin{tabular}{@{}ll rrrr rrrr@{}}
\toprule
& & \multicolumn{4}{c}{\textbf{GPT-5.4}} & \multicolumn{4}{c}{\textbf{SwissBERT}} \\
\cmidrule(lr){3-6} \cmidrule(lr){7-10}
\textbf{Lang.} & \textbf{Label} & \textbf{Prec.} & \textbf{Rec.} & \textbf{F1} & \textbf{Macro F1}
              & \textbf{Prec.} & \textbf{Rec.} & \textbf{F1} & \textbf{Macro F1} \\
\midrule
\multirow{3}{*}{DE}
  & Claim    & 0.395 & 0.447 & 0.419 & \multirow{3}{*}{0.543} & 0.871 & 0.704 & \textbf{0.778} & \multirow{3}{*}{0.866} \\
  & Premise  & 0.807 & 0.614 & 0.698 & & 0.873 & 0.942 & \textbf{0.906} & \\
  & Non-Arg. & 0.379 & 0.792 & 0.512 & & 0.936 & 0.889 & \textbf{0.912} & \\
\midrule
\multirow{3}{*}{FR}
  & Claim    & 0.381 & 0.423 & 0.400 & \multirow{3}{*}{0.528} & 0.865 & 0.696 & \textbf{0.771} & \multirow{3}{*}{0.856} \\
  & Premise  & 0.810 & 0.604 & 0.692 & & 0.866 & 0.932 & \textbf{0.898} & \\
  & Non-Arg. & 0.355 & 0.795 & 0.491 & & 0.905 & 0.893 & \textbf{0.899} & \\
\midrule
\multirow{3}{*}{IT}
  & Claim    & 0.432 & 0.478 & 0.454 & \multirow{3}{*}{0.572} & 0.870 & 0.695 & \textbf{0.773} & \multirow{3}{*}{0.853} \\
  & Premise  & 0.816 & 0.647 & 0.722 & & 0.867 & 0.936 & \textbf{0.901} & \\
  & Non-Arg. & 0.897 & 0.876 & \textbf{0.886} & & 0.897 & 0.876 & \textbf{0.886} & \\
\midrule
\multirow{3}{*}{RM}
  & Claim    & 0.435 & 0.496 & 0.464 & \multirow{3}{*}{0.582} & 0.864 & 0.702 & \textbf{0.775} & \multirow{3}{*}{0.862} \\
  & Premise  & 0.812 & 0.655 & 0.725 & & 0.868 & 0.936 & \textbf{0.901} & \\
  & Non-Arg. & 0.435 & 0.779 & 0.558 & & 0.930 & 0.889 & \textbf{0.909} & \\
\midrule
\multirow{3}{*}{Overall}
  & Claim    & 0.411 & 0.461 & 0.434 & \multirow{3}{*}{0.556} & 0.867 & 0.699 & \textbf{0.774} & \multirow{3}{*}{0.859} \\
  & Premise  & 0.811 & 0.630 & 0.709 & & 0.869 & 0.937 & \textbf{0.901} & \\
  & Non-Arg. & 0.392 & 0.792 & 0.524 & & 0.917 & 0.887 & \textbf{0.902} & \\
\bottomrule
\end{tabular}
\caption{Component classification results per language and overall for GPT-5.4 and SwissBERT.
Precision, Recall, F1 = per-class F1, and Macro F1
reported per language.}
\label{tab:component-comparison}
\end{table*}

\paragraph{Relation and target prediction results.}
We evaluate relation and target prediction jointly as a single label (e.g., 
\textit{Support\_Claim~4}) on all segments with argumentative ground truth 
(i.e., \textit{claim} or \textit{premise}), and report results over correctly classified 
components (\textit{correct comp.}). Since SwissBERT's performance is stronger on the component classification, the model classified more instances in this second subtask.
Results are reported in Table~\ref{tab:gpt-relation}.

For both models, performance on this subtask is substantially lower than for component classification. SwissBERT's performance remains stable across languages and consistently outperforms GPT, likely due to its language-specific adaptations and training on Swiss language varieties as well as fine-tuning on our dataset. In contrast, zero-shot GPT shows greater variation across languages: performance on French is lowest (macro F1: 0.233), while Italian and Romansh perform comparably and most strongly (macro F1: 0.249 and 0.359--0.361, respectively).

When restricting evaluation to correctly classified components, macro F1 is at 0.315 and exact match at 0.427 for GPT, and 0.477 and 0.489 respectively
for SwissBERT. This isolates the intrinsic difficulty of relation prediction from the compounding effect of component classification errors.


\begin{table}[h]
\centering
\small
\setlength{\tabcolsep}{6pt}
\begin{tabular}{l rrr rrr}
\toprule
& \multicolumn{3}{c}{\textbf{GPT-5.4}}   & \multicolumn{3}{c}{\textbf{SwissBERT}} \\
\cmidrule(lr){2-4} \cmidrule(lr){5-7}
\textbf{Lang.} & \textbf{N} & \textbf{EM} & \textbf{F1} & \textbf{N} & \textbf{EM} & \textbf{F1}\\
\midrule
DE      & 1455 & 0.40 & 0.30 & 2244 & 0.51 & 0.59 \\
FR      & 1417 & 0.34 & 0.23 & 2220 & 0.48 & 0.47 \\
IT      & 1538 & 0.47 & 0.36 & 2226 & 0.48 & 0.47 \\
RM      & 1567 & 0.49 & 0.36 & 2232 & 0.49 & 0.48 \\
\midrule
Overall & 5977 & 0.43 & 0.32 & 8922 & 0.489 & 0.477 \\
\bottomrule
\end{tabular}
\caption{GPT 5.4 and SwissBERT relation and target prediction results evaluated only on units where the component label was correctly predicted. EM = exact match on the joint relation+target label. F1 = Macro F1 Score.}
\label{tab:gpt-relation}
\end{table}

\paragraph{Discussion.} 
The most consistent pattern across all languages and both models is the difficulty of
distinguishing \textit{claims} from \textit{premises}. For GPT, 29--35\% of true 
\textit{claims} are predicted as \textit{premises}, and 26--29\% of true \textit{premises} are 
predicted as \textit{claims}. SwissBERT exhibits a similar asymmetry: while roughly 
30\% of Claims are misclassified as Premises, the reverse confusion is much 
less frequent (only 4.4\% of Premises are predicted as Claims), reflecting 
the class imbalance in favor of Premises. 
This mirrors the dominant human disagreement pattern identified in 
Section~\ref{sec:iaa}, where Claim--Premise confusion was the most frequent 
subtype (accounting for 57--70\% of all component disagreements across 
languages).

Both models show strong performance on the \textit{non-Argumentative} class. 
For GPT, recall is consistently above 0.779, indicating that non-argumentative  segments are reliably detected, although the relatively low precision (0.355--0.435) suggests that some argumentative segments, particularly short or implicit ones, are incorrectly labeled as non-argumentative. SwissBERT, in contrast, identifies non-argumentative segments more reliably overall, with 88.7\% correct classifications. Misclassifications of non-argumentative segments are relatively infrequent and predominantly skewed toward the \textit{premise} class (9.0\%), while confusion with \textit{claim} is rare 
(2.3\%). Conversely, argumentative segments are almost never misclassified as non-argumentative (approximately 1\% for both \textit{claims} and \textit{premises}).

The relation+target task remains an open challenge even under oracle 
component labels, that is, only evaluating relation+target labels if the component prediction is correct, (exact match $\leq$ 0.485), highlighting that identifying 
the correct \textit{direction} of argumentation and the precise \textit{target} 
segment requires a deeper understanding of document structure than GPT-5.4 
currently exhibits in a zero-shot setting. Despite SwissBERT's stronger 
component classification performance, it faces similar difficulties in this 
subtask, suggesting that relation prediction constitutes a distinct challenge 
beyond component identification.

\section{Conclusion}
We presented \texttt{Swiss4Arg}, a multilingual, high-quality argument mining dataset based on Swiss federal voting brochures, covering all four national languages including Romansh, a low-resource language underrepresented in NLP. Beyond introducing a new benchmark, the dataset provides a testbed for studying multilingual argument mining in parallel documents. 
Our analyses demonstrate both the potential and the remaining challenges of multilingual argument mining. While annotation agreement is largely consistent across languages, distinguishing claims from premises and predicting argument relations remain difficult for both human annotators and strong baseline models. We hope that \texttt{Swiss4Arg} will foster future work on multilingual and low-resource AM, as well as on more robust cross-lingual modeling approaches.


\section*{Limitations}
Although \texttt{Swiss4Arg} covers all four national Swiss languages, we were limited to three annotators per language. Recruiting qualified native annotators for Romansh is particularly challenging due to its small speaker community. Therefore, assembling larger annotation pools is difficult. A higher number of annotators would likely yield a more robust gold standard, especially for borderline cases requiring adjudication but would be time-consuming and costly.

The dataset comprises 53 federal voting issues, constrained by the availability of structured digital brochures through the VoteInfo API. 

Finally, our experimental evaluation is limited to zero-shot prompting with GPT-5.4 and monolingual supervised baselines. We do not investigate few-shot prompting or cross-lingual transfer, despite the parallel structure of \texttt{Swiss4Arg} making such experiments an avenue for future work.

\bibliography{custom}

\appendix
\nolinenumbers
\section{Example Text (German)}
\label{app:text}
\begin{figure*}
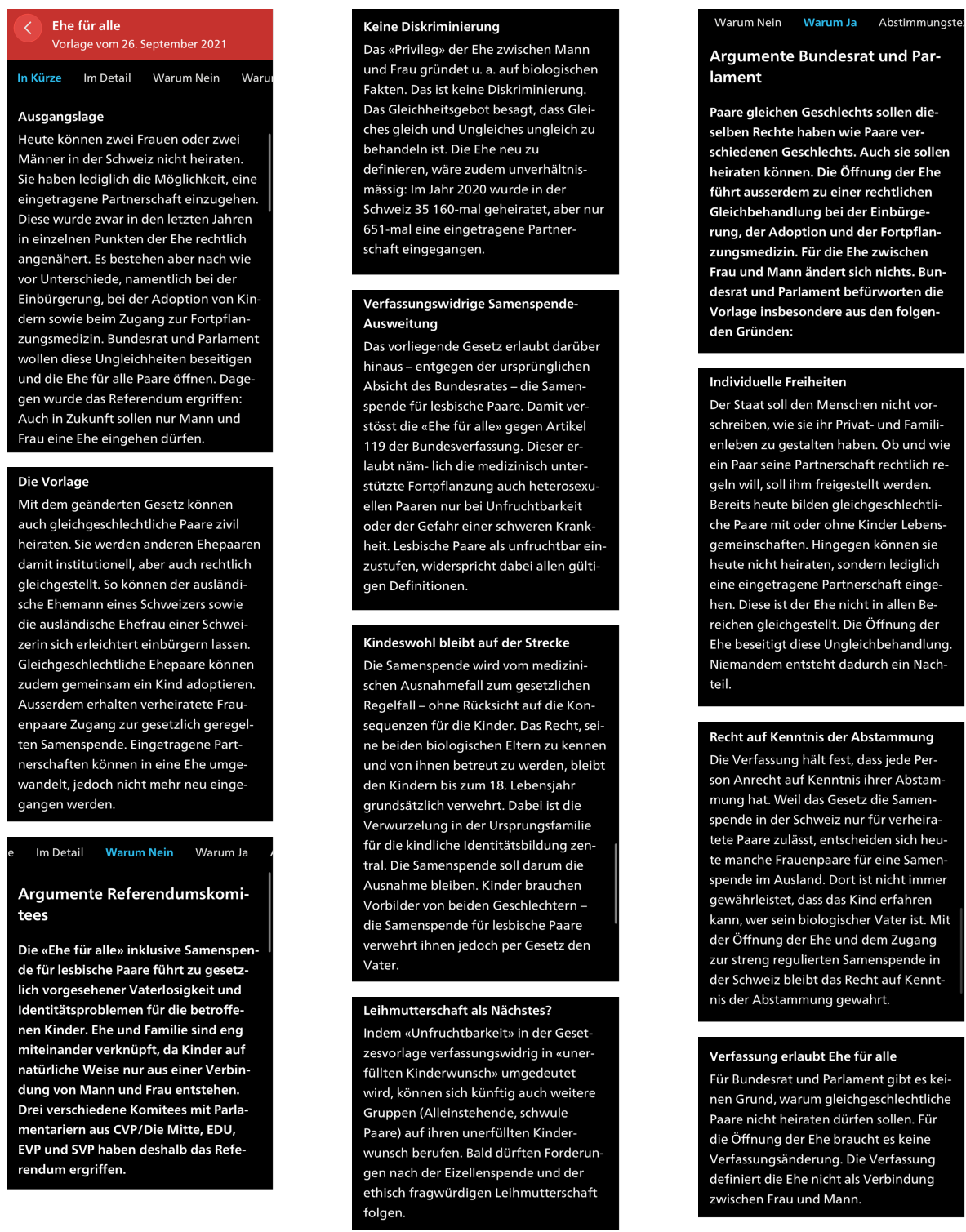

    \centering
    \begin{minipage}[t]{0.3\textwidth}
        \strut\vspace{-\baselineskip}\par
        \centering
        \includesvg[width=0.9\linewidth]{final_images/Picture1}\par\vspace{2mm}
        \includesvg[width=0.9\linewidth]{final_images/Picture2}\par\vspace{2mm}
        \includesvg[width=0.9\linewidth]{final_images/Picture3}\par\vspace{2mm}
    \end{minipage}
    \hfill
    \begin{minipage}[t]{0.3\textwidth}
        \strut\vspace{-\baselineskip}\par
        \centering
                \includesvg[width=0.9\linewidth]{final_images/Picture4}\par\vspace{2mm}
        \includesvg[width=0.9\linewidth]{final_images/Picture5}\par\vspace{2mm}
        \includesvg[width=0.9\linewidth]{final_images/Picture6}\par\vspace{2mm}
        \includesvg[width=0.9\linewidth]{final_images/Picture7}

    \end{minipage}
    \hfill
    \begin{minipage}[t]{0.3\textwidth}
        \strut\vspace{-\baselineskip}\par
        \centering
                \includesvg[width=0.9\linewidth]{final_images/Picture8}\par\vspace{2mm}
        \includesvg[width=0.9\linewidth]{final_images/Picture9}\par\vspace{2mm}
        \includesvg[width=0.9\linewidth]{final_images/Picture11}\par\vspace{2mm}
        \includesvg[width=0.9\linewidth]{final_images/Picture13}
    \end{minipage}
    \caption{Example text from VoteInfo App. More examples and the texts in the other three languages can be found there or at \url{https://www.bk.admin.ch/de/sammlung-der-abstimmungsbuechlein-seit-1978}}
    \label{fig:placeholder}
\end{figure*}

\newpage
\section{Distribution of Pro \& Con sections in the Brochures}

\begin{strip}
    \centering
    \includesvg[width=\textwidth]{final_images/pro_con_diverging}
    \captionof{figure}{Distribution of pro and con spans by text. Numbers on the x-axis correspond to text ids as assigned in the database behind VoteInfo.}
    \label{fig:pro_con_distrib}
\end{strip}

\newpage
\section{Major Claim Generation Prompt}\label{claim_prompt}
\begin{quote}
\small
SYSTEM\_PROMPT = You are an assistant specialized in summarizing political and legal texts.

USER\_PROMPT\_TEMPLATE = \{text\}

                        State the core claim of the initiative as a single, neutral,
                        normative sentence in German.

                        The sentence must:
                        - express a proposal that can be voted for or against
                        - use prescriptive language with "soll" or similar wording
                        - state what ought to be changed by the initiative
                        - contain no introductory phrases or references to the initiative
                        - contain no arguments, explanations, or additional details

                        Return ONLY valid JSON with exactly this structure:
                        {{
                        ''major\_claim": ''..."
                        }}

\end{quote}

\section{Disagreement within languages}\label{per_lang_disagreement}
Table~\ref{tab:adjudication2} breaks down the frequency of within-language
disagreements, reflecting annotation difficulty per language independently
of the cross-lingual adjudication process. Italian shows the highest rate
of internal disagreement, while French and Romansh are the
most internally consistent.

\begin{table}[H]
\centering
\small
\begin{tabular}{lrrr}
\toprule
\textbf{Language} & \textbf{2-vs-1} & \textbf{3-way} & \textbf{\% Majority} \\
\midrule
French   & 811  & 31  & 96.3\% \\
German   & 769  & 52  & 93.7\% \\
Italian  & 1387 & 131 & 91.4\% \\
Romansh  & 943  & 44  & 95.5\% \\
\midrule
Overall  & 3910 & 258 & 93.8\% \\
\bottomrule
\end{tabular}
\caption{Within-language disagreement types per language team.
2-vs-1: two annotators agree, one disagrees.
3-way: all three annotators assign different labels.
The \% Majority column reflects the proportion of within-language
disagreements resolvable by majority vote and serves as a proxy
for annotation difficulty per language.}
\label{tab:adjudication2}
\end{table}

Table~\ref{tab:krip} summarizes the resulting agreement scores. French and 
German achieve the highest reliability, approaching or exceeding the commonly 
accepted threshold of $\alpha = 0.67$ for both annotation tasks. Agreement 
is lower for Romansh, particularly for relation annotation, while Italian 
exhibits substantially lower agreement on both dimensions, driven by one 
annotator whose annotations diverge from those of both teammates.

\begin{table}[H]
\centering
\small
\setlength{\tabcolsep}{6pt}
\begin{tabular}{lcc}
\toprule
\textbf{Language} & \textbf{Component} & \textbf{Rel.+Target} \\
\midrule
French   & 0.662 & 0.693 \\
German   & 0.685 & 0.678 \\
Romansh  & 0.624 & 0.581 \\
Italian  & 0.463 & 0.371 \\
\bottomrule
\end{tabular}
\caption{Pooled Krippendorff's $\alpha$ per language for component 
and relation+target annotation.}
\label{tab:krip}
\end{table}

\newpage
\section{Annotations Confusion Matrices}\label{app:cms}
\vspace{10mm}
\begin{figure}[h!]
    \centering
    \includesvg[width=\linewidth]{final_images/fr}
    \includesvg[width=\linewidth]{final_images/fr2}
    \includesvg[width=\linewidth]{final_images/fr3}
    \caption{Pairwise confusion matrices for component labels within the 
    French annotation team.}
    \label{fig:conf_french}
\end{figure}

\begin{figure}
    \centering
    \includesvg[width=\linewidth]{final_images/gm}
    \includesvg[width=\linewidth]{final_images/gm2}
    \includesvg[width=\linewidth]{final_images/gm3}
    \caption{Pairwise confusion matrices for component labels within the 
    German annotation team.}
    \label{fig:conf_german}
\end{figure}
\newpage
\begin{figure}[h!]
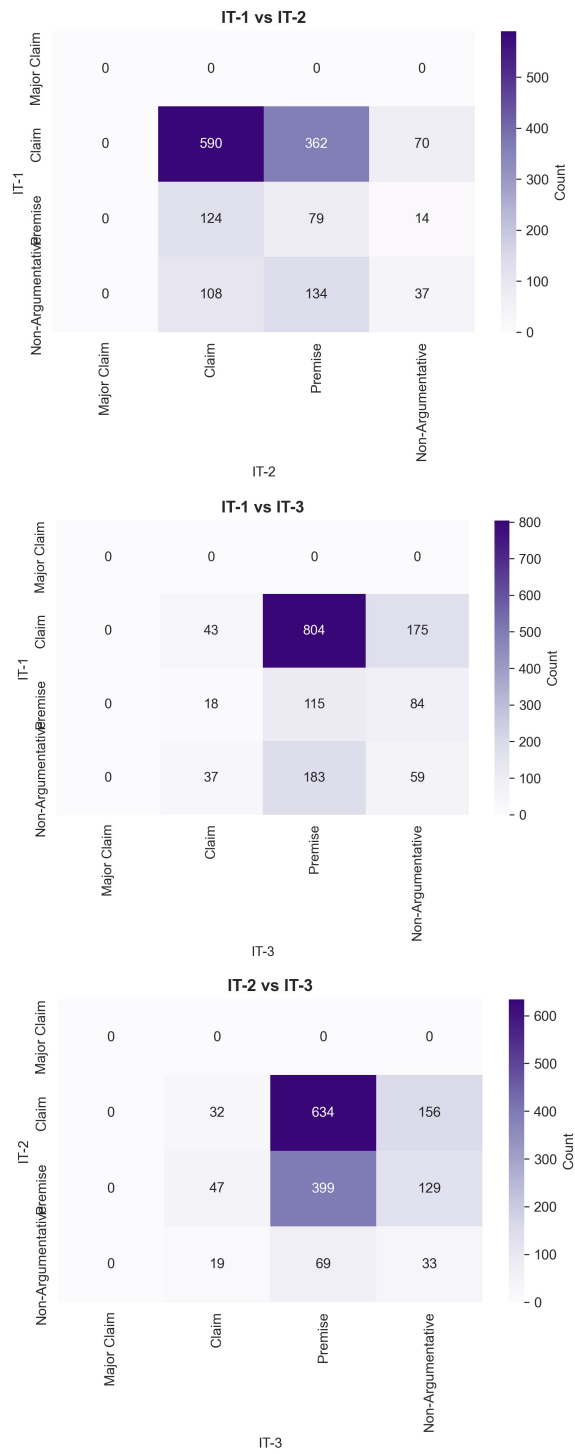

    \centering
    \includesvg[width=\linewidth]{final_images/it}
    \includesvg[width=\linewidth]{final_images/it2}
    \includesvg[width=\linewidth]{final_images/it3}
    \caption{Pairwise confusion matrices for component labels within the 
    Italian annotation team.}
    \label{fig:conf_italian}
\end{figure}
\begin{figure}[H]
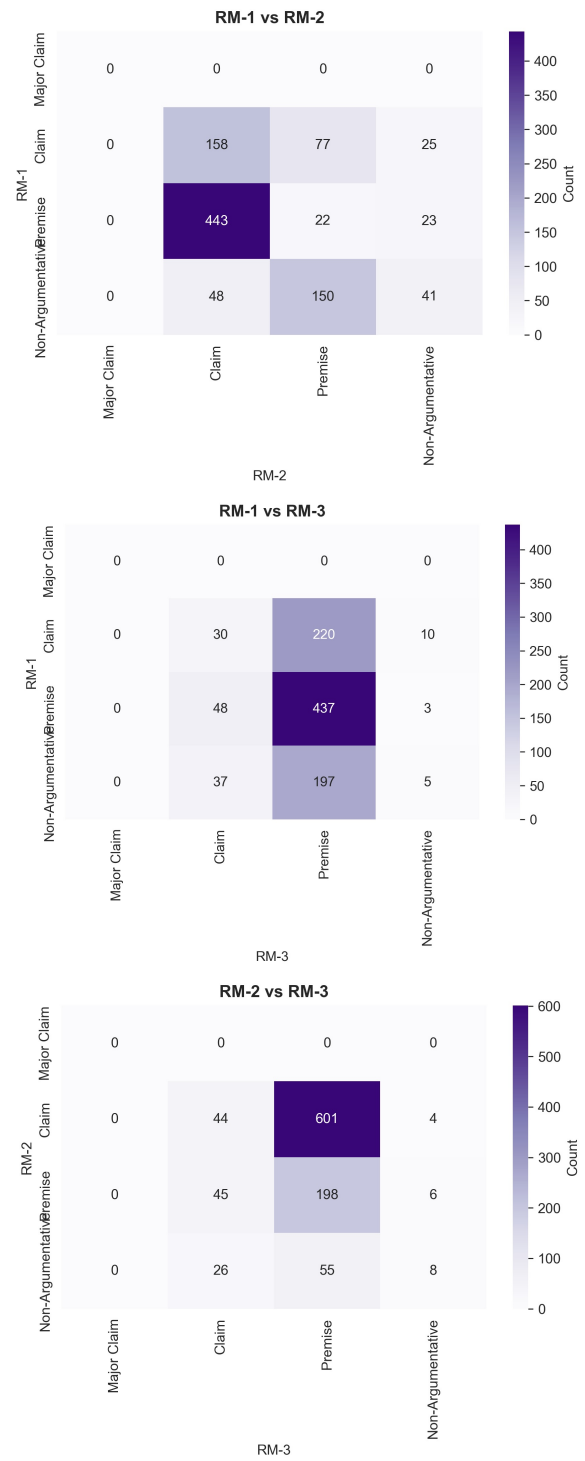

    \centering
    \includesvg[width=\linewidth]{final_images/rm}
    \includesvg[width=\linewidth]{final_images/rm2}
    \includesvg[width=\linewidth]{final_images/rm3}
    \caption{Pairwise confusion matrices for component labels within the 
    Romansh annotation team.}
    \label{fig:conf_romansh}
\end{figure}

\newpage
\section{GPT-5.4 Annotation Prompts}
\label{app:prompts}
\begin{quote}
\small
\textbf{SYSTEM PROMPT}

You are an expert in computational argumentation and argumentation mining, trained in the theoretical frameworks of Stab \& Gurvych (2017) and Peldszus \& Stede (2016). You are specialized in annotating Swiss federal vote brochures for argumentation structure in \{language\}. These brochures are official documents presenting PRO and CON arguments for political initiatives.

Your task is to label each text segment with a component type and, where applicable, a relation and target.

\textbf{Theoretical Framework}

Following Stab \& Gurevych (2017), argumentative discourse consists of:

\textbf{Component Types}
\begin{itemize}
  \item \textbf{major\_claim} -- The central thesis of the entire brochure. One per document. This is already provided. Relate all ``claims'' to it. Do not annotate anything else as major\_claim.
  \item \textbf{claim} -- A debatable sub-position that directly supports or attacks the major claim. States an opinion or evaluation, provides no evidence on its own.
  \item \textbf{premise} -- A reason, fact, or evidence that supports or attacks a specific claim. Contains data, statistics, examples, consequences, or law references. Tip: if the segment contains a number, a named example, or a statistic, it is almost certainly a premise.
  \item \textbf{non-argumentative} -- Background, procedural text, structural headers, or neutral descriptions. This includes colon-ending introductions, and sentences describing the legal situation without taking a position.
\end{itemize}

\textbf{Relations}

Every claim and every premise must have exactly one relation and one target:
\begin{itemize}
  \item claim $\rightarrow$ supports OR attacks $\rightarrow$ major\_claim
  \item premise $\rightarrow$ supports OR attacks $\rightarrow$ [specific claim by segment\_id]
\end{itemize}

Non-argumentative segments and the major claim have no relation and no target.
\begin{itemize}
  \item \textbf{supports}: the unit reinforces, justifies, or provides evidence for its target.
  \item \textbf{attacks}: the unit contradicts, undermines, or opposes its target.
\end{itemize}

\textbf{Strict Structural Rules}
\begin{itemize}
  \item Claims NEVER point to other claims
  \item Premises NEVER point to the major claim directly
  \item Premises NEVER point to other premises
  \item The hierarchy is always: premise $\rightarrow$ claim $\rightarrow$ major\_claim
\end{itemize}
\end{quote}

\begin{quote}
\small
\textbf{USER PROMPT TEMPLATE}

Annotate every segment below. This is the major claim you should relate all claims to.

MAJOR CLAIM (segment\_id: 1): \{major\_claim\_text\}

SEGMENTS:
\{segments\}

For each segment return exactly:
\begin{itemize}
  \item segment\_id : integer, from the input
  \item component : one of claim / premise / non-argumentative
  \item relation : one of supports / attacks / null
  \item target : segment\_id as integer, or ``major\_claim'', or null
\end{itemize}

Rules:
\begin{itemize}
  \item relation and target must be null for non-argumentative and major\_claim
  \item claim target must always be ``major\_claim'' (the string, not a number)
  \item premise target must be the integer segment\_id of the claim it supports or attacks
  \item annotate every segment, no skipping
\end{itemize}

Return only valid JSON, no prose:

\{ ``annotations": [ \{``segment\_id": 2, ``component": ``claim", ``relation": ``supports", ``target": ``major\_claim"\}, \{``segment\_id": 3, ``component": ``premise", ``relation": ``attacks", ``target": 2\}, \{``segment\_id": 4, ``component": ``non-argumentative", ``relation": null, ``target": null\} ] \}
\end{quote}

\section{SwissBERT Hyperparameters}\label{app:swissbert_hyperparams}
We tuned the hyperparameters \textit{learning rate} (1e-05, 2e-05, 5e-05) and \textit{number of epochs} (3, 5, and 8). We found that for component prediction learning rate = 2e-05 and 8 epochs, and for relation+target prediction learning rate = 5e-05 and 8 epochs yielded the best results. That is, a macro F1 score of 0.87 and 0.48, respectively for the tasks. 

\newpage


\section{Component Distribution per Language}\label{per_lang_labels}
\begin{figure}[h]
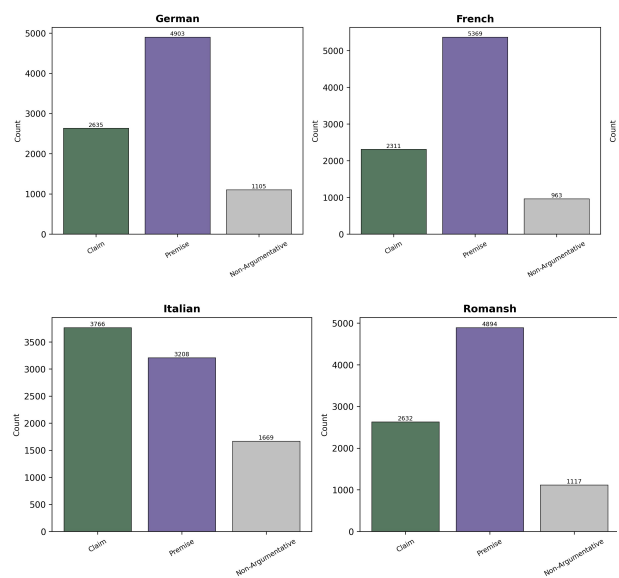

    \includesvg[width=0.5\textwidth]{final_images/per_language_component1} \\
    \includesvg[width=0.5\textwidth]{final_images/per_language_component2}
    \captionof{figure}{Component label distribution per language across all individual annotators. Italian annotators assign Claim at a substantially higher rate (42.0\%) and Premise at a lower rate (35.8\%) compared to the other 
language teams (Claim: 25.8--29.4\%, Premise: 54.6--59.9\%), driven 
primarily by IT-1.}
    \label{fig:per_lang_dist}
\end{figure}

\section{GPT-5.4 Confusion Matrices}\label{app:gpt_cms}
\begin{figure}[h!]
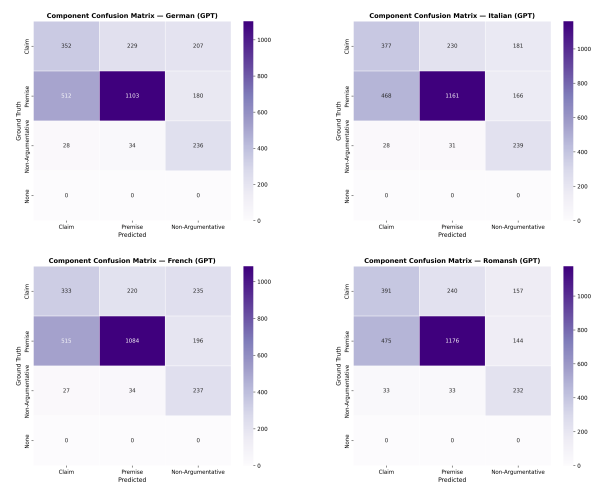

    \centering
    \begin{minipage}[t]{0.45\linewidth}
        \centering
        \includesvg[width=\linewidth]{final_images/confusion_matrix_german}\par\vspace{4pt}
        \includesvg[width=\linewidth]{final_images/confusion_matrix_french}
    \end{minipage}
    \hfill
    \begin{minipage}[t]{0.45\linewidth}
        \centering
        \includesvg[width=\linewidth]{final_images/confusion_matrix_italian}\par\vspace{4pt}
        \includesvg[width=\linewidth]{final_images/confusion_matrix_romansh}
    \end{minipage}
    \caption{Component confusion matrices for GPT-5.4 predictions vs.\ 
    majority-vote ground truth, per language. Rows = ground truth, 
    columns = predicted. Major Claims are excluded.}
    \label{fig:gpt-confusion}
\end{figure}


\end{document}